\documentclass{article}

\usepackage[preprint]{neurips_2026}

\usepackage{hyperref}
\usepackage{url}
\usepackage{amsfonts}       
\usepackage{nicefrac}       
\usepackage{microtype}      
\usepackage{xcolor}         
\usepackage{amsmath, amssymb, amsthm}
\usepackage{algorithm, algpseudocode}
\usepackage{graphicx}
\usepackage{booktabs}

\title{Skeletons in Flow: Graph Structured Flow Matching for Human Motion Prediction\thanks{This research is based on work supported in part by AFOSR grant FA9550-19-1-0169, and Office of Naval Research Grant N00014-13-1-0151.
Any opinions, findings and conclusions or recommendations expressed in this material are those of the author(s) and do not necessarily reflect the views of the sponsoring agency.}}

\author{%
  Yixuan Wang \thanks{corresponding author.}\quad Brandon C. Fallin\\
  Department of Mechanical and Aerospace Engineering\\ University of Florida\\
  \texttt{\{wang.yixuan, brandonfallin\}@ufl.edu}\\
  \And Warren E. Dixon\\
  Department of Engineering\\
  Virginia Tech\\
  \texttt{wdixon@vt.edu}
}

\begin{document}

\maketitle

\begin{abstract}
Human motion prediction requires diverse future trajectories that remain consistent with observed motion and the articulated physical structure of the body. Skeletal constraints restrict individual poses, while coordinated motion depends on spatial interactions (between connected joints) and temporal interactions (between  time instants). To facilitate human motion prediction in light of these constraints and interactions, we introduce Graph Structured Flow Matching (GSFM), which transports the complete future skeletal trajectory through a single conditional velocity field. The trajectory produces a spatiotemporal skeleton graph, and spatial and temporal attention couple its evolution according to skeletal relations and physical time offsets. Bone directions lie on unit spheres relative to a root joint, and tangent evolution preserves input bone lengths throughout generation. We train a learned velocity field through conditional flow matching along geodesic paths connecting random trajectories centered on the last-observed pose to recorded future trajectories. Experiments on the Archive of Motion capture As Surface Shapes (AMASS) dataset evaluate prediction accuracy, diversity calibration, and motion statistics. We demonstrate the contributions of spatial and temporal message passing in the developed architecture through an ablation study.
%
%
GSFM models trained on AMASS also perform competitively on the Human3.6M skeleton without parameter updates or retraining, demonstrating applicability to an unseen skeletal structure.
\end{abstract}

\section{Introduction}
\label{sec:introduction}
Human motion prediction aims to infer future pose sequences from observed motion.
The same observed pose sequence can admit multiple plausible continuations.
To this end, stochastic prediction is used to model a conditional distribution rather than a single trajectory~\citep{yuan2020dlow,barquero2023belfusion}.
To effectively predict future motion, the modeled conditional distribution must capture (i) the constraints induced by the physical connections of joints, and (ii) the possibility of many potential future trajectories from a single observed pose.
Within a single discrete time frame, joints are coupled spatially and are limited in their state through prescribed bone lengths.
Temporally, future joint behavior can be inferred from the previously observed states.
As such, the predicted future skeletal state must demonstrate physically plausible temporal motion while simultaneously preserving the physical constraints of the skeletal geometry~\citep{barquero2023belfusion,curreli2025nonisotropic}.

Recently, generative approaches to human motion prediction model future trajectories through latent variable models and diffusion processes \citep{yuan2020dlow,barquero2023belfusion,chen2023humanmac,sun2024comusion}.
In the physical domain, geometric representations of the skeletal structure and tailored network architectures aim to emphasize links between physical joints~\citep{curreli2025nonisotropic,curreli2026equifusion}.
In the temporal domain, flow matching represents a promising method for generating plausible future trajectories given past, observed behavior.
Specifically, flow matching learns a continuous velocity field to transport data from an initial, prior distribution to the true data distribution~\citep{lipman2023flow,chen2024flow}.
Riemannian flow matching \citep{chen2024flow} generates samples on manifolds and has recently been applied to text-conditioned human motion generation on a product of a translation factor and unit-quaternion spheres~\citep{miao2026riemannian}.
These developments motivate the joint consideration of physical constraints and temporal evolution when predicting human motion.
However, while geometrically constrained generation limits outputs to physically allowable configurations, this factor alone does not ensure that the resulting predicted trajectories are plausible and diverse.
This leads to the central question which we consider in this work: \textbf{Can the entire future skeletal trajectory be processed as a single data point through a learned flow which (i) couples spatial and temporal components, and (ii) preserves the input skeletal geometry?}

\begin{figure}[t]
\centering
\includegraphics[width=\textwidth]{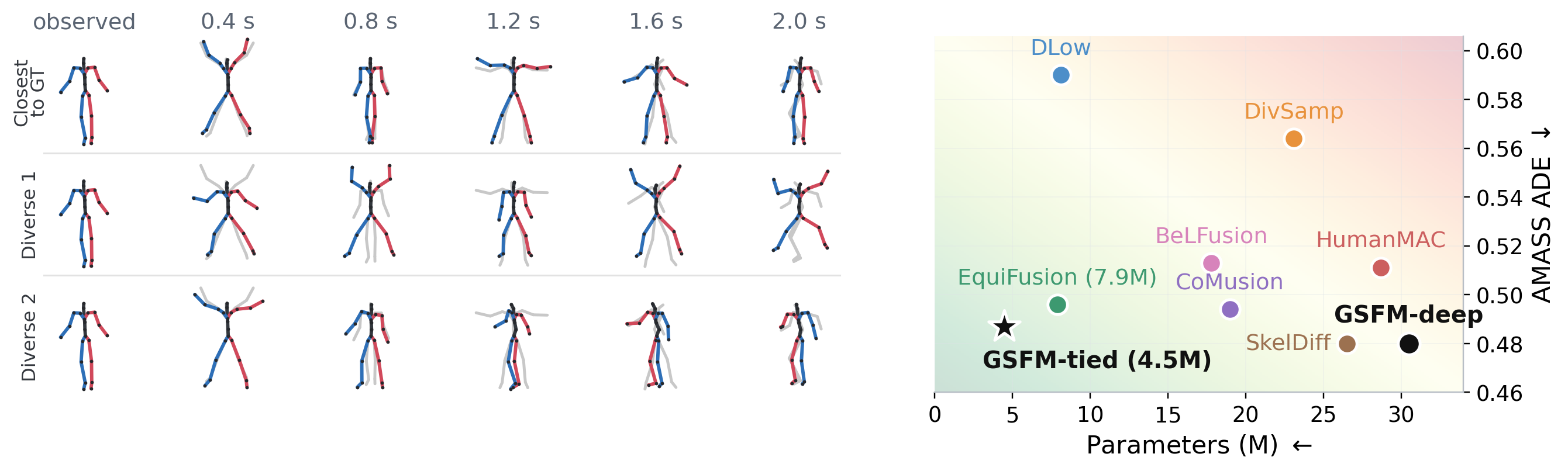}
\caption{\textbf{Graph structured flow matching for human motion prediction.}
Left: the sample with the lowest average displacement error (ADE) and two diverse samples selected by farthest point sampling among 50 generated futures for an AMASS test segment.
%
%
The observed column shows the last pose of the $0.5 s$ input window, and gray poses denote the
ground truth.
Right: AMASS ADE against single-instance parameter counts.
Baseline architectures and their parameter counts are sourced from \citet{curreli2026equifusion}.
%
%
The GSFM-tied configuration reaches near-best precision with $4.5 \mathrm{M}$ parameters, while the GSFM-deep configuration matches the best published ADE.}
\label{fig:teaser}
\end{figure}

To this end, we develop a \emph{Graph Structured Flow Matching} (GSFM) model.
Our central insight is to make the complete future skeleton trajectory the object transported by the flow.
We represent input trajectories as spatiotemporal skeletal graphs and learn a single conditional velocity field that jointly evolves all future node states.
%
%
Conditioned on the observed motion, the flow operates on the entire candidate future trajectory, rather than advancing a single pose through physical time.
Our representation of the skeleton's pose and its tangent flow preserve the physical constraints (bone lengths and connections) of the input trajectory throughout this evolution.
Figure~\ref{fig:teaser} demonstrates how our model generates multiple, alternative future trajectories from the same input observation.

This paper contributes in the following three main aspects:
\begin{enumerate}
    \item \textbf{Graph structured conditional transport.}
    We formulate human motion prediction as the joint evolution of a complete future spatiotemporal skeletal trajectory under a single conditional velocity field.
    \item \textbf{Skeletal parameterization of flow.}
    We incorporate spatial and temporal relations into the velocity field and preserve prescribed bone lengths throughout generation, with the skeleton structure supplied as an input.
    \item \textbf{Evaluation on AMASS} and transfer to unseen data.
    We evaluate prediction quality, diversity calibration, and motion statistics through AMASS prediction, spatial and temporal ablations, and zero-shot transfer to the Human3.6M skeleton.
\end{enumerate}

\noindent\textbf{Notation.}
Let $\mathbb N_{+}$ denote the positive integers.
For $n\in\mathbb N_{+}$, let $\mathbb R^n$ denote Euclidean space with dimension $n$, and let $I_n$ denote the identity matrix of size $n$.
For $a,b\in\mathbb R^n$, the Euclidean inner product and its induced norm are $\langle a,b\rangle\triangleq a^{\mathsf T}b$ and $\|a\|\triangleq\sqrt{\langle a,a\rangle}$, respectively.
Let $\lvert\mathcal A\rvert$ denote the cardinality of the finite set $\mathcal{A}$.
For random vectors $W$ and $C$, let $\mathbb{E}[W]$ and $\mathbb{E} [W\mid C]$ denote the expectation and conditional expectation, respectively, provided that $W$ is integrable. 
%
%
We denote the standard Gaussian distribution on $\mathbb{R}^n$ by $\mathcal N(0,I_n)$.
Lastly, let the symbol $\triangleq$ denote a definition.

\section{Related Work}
\label{sec:related_work}
%
Latent generative models and specialized sampling strategies have been used to represent multiple possible motion futures \citep{walker2017pose,mao2021generating,dang2022diverse}.
The Diversifying Latent Flows (DLow)~\citep{yuan2020dlow} approach leverages a pre-trained generative model, and introduces a novel sampling method which emphasizes the generation of diverse future trajectories.
Human MAsked motion Completion (HumanMAC \citep{chen2023humanmac}) formulates prediction as masked motion completion and incorporates observed motion during diffusion sampling in the discrete cosine transform domain.
CoMusion \citep{sun2024comusion} combines transformer reconstruction of corrupted motion with graph convolutional refinement conditioned on the observed motion within a diffusion framework.
BeLFusion \citep{barquero2023belfusion} performs diffusion in a latent behavior representation and couples sampled behaviors with the observed motion.
SkeletonDiffusion \citep{curreli2025nonisotropic} incorporates skeletal structure into latent diffusion through a nonisotropic Gaussian formulation.
EquiFusion~\citep{curreli2026equifusion} removes the need to hard-code a fixed skeleton, as its permutation-equivariant latent diffusion model takes the kinematic tree as an explicit input, which enables zero-shot prediction on unseen skeletons.
Additional related work is discussed in Appendix~\ref{app:additional_related}.

\section{Preliminaries}
\label{sec:preliminaries}

Throughout the manuscript, we use the following geometric operations to construct the training paths and integrate the learned velocity field. We refer the reader to~\citet{lee2018riemannian} and~\citet{chen2024flow} for a deeper exposition of these concepts.
First, consider the unit sphere $\mathbb S^2\triangleq\{p\in\mathbb R^3:\|p\|=1\}$, equipped with the metric induced by the Euclidean inner product.
For a point $p\in\mathbb S^2$, let $\mathcal{T}_p\mathbb S^2$ denote the tangent space (i.e., the plane orthogonal to $p$).
For $w\in\mathbb R^3$, the orthogonal projection $\Pi_p(w)$ onto this tangent space is defined as
\begin{align*}
\mathcal{T}_p\mathbb S^2\triangleq\{w\in\mathbb R^3:\langle p,w\rangle=0\},\quad \Pi_p(w)\triangleq w-\langle p,w\rangle p.
\end{align*}
For $q\in\mathbb S^2\setminus\{p,-p\}$, the geodesic distance is $\rho(p,q)\triangleq\arccos\langle p,q\rangle$.
For nonzero $u\in\mathcal T_p\mathbb S^2$, the exponential map gives a point on the sphere, while the logarithm map returns the initial tangent vector of the minimizing geodesic from $p$ to $q$.
The exponential map and logarithm map are defined as
\begin{align*}
\operatorname{\textbf{exp}}_p(u)\triangleq\cos(\|u\|)p+\frac{\sin(\|u\|)}{\|u\|}u,\quad \operatorname{\textbf{log}}_p(q)\triangleq\frac{\rho(p,q)}{\sin\rho(p,q)}\big(q-\langle p,q\rangle p\big),
\end{align*}
respectively.
The continuous extensions of the exponential and logarithm maps satisfy $\operatorname{\textbf{exp}}_p(0)=p$ and $\operatorname{\textbf{log}}_p(p)=0$, respectively.
The antipodal point $-p$ is excluded because the minimizing geodesic is not unique.
For $q\in\mathbb S^2\setminus\{-p\}$, parallel transport along the minimizing geodesic is the map $\mathcal T_{p\to q}:\mathcal T_p\mathbb S^2\to\mathcal T_q\mathbb S^2$.
For $s\in[0,1]$, define the geodesic $\gamma_s\triangleq\operatorname{\textbf{exp}}_p(s\operatorname{\textbf{log}}_p(q))$.
For $u\in\mathcal T_p\mathbb S^2$, parallel transport and the geodesic velocity satisfy the following expressions:
\begin{align*}
\mathcal T_{p\to q}(u)\triangleq u-\frac{\langle q,u\rangle}{1+\langle p,q\rangle}(p+q),\quad \frac{d\gamma_s}{ds}=\mathcal T_{p\to\gamma_s}\!\left(\mathbf{log}_p(q)\right).
\end{align*}
Parallel transport preserves the norm, which yields $\|\mathcal T_{p\to q}(u)\|=\|u\|$.
On a Euclidean factor of a product manifold, the tangent space is the Euclidean space itself, the exponential map is addition, the logarithm map is subtraction, and projection and parallel transport are the identity.
On a finite product of Euclidean and spherical factors with the product metric, tangent spaces form the corresponding product and these four operations (exponential map, logarithm map, projection, and parallel transport) act separately on each factor.

\section{Problem Setting}
\label{sec:problem_setting}
In this work, we consider the conditional generation of future human motion from a sequence of observed poses.
The skeleton contains $J$ joints, where $J\in\mathbb N_+$.
The joint set is $V\triangleq\{0,\ldots,J-1\}$, where joint $0$ denotes the root.
The parent map $\mathsf{p}:V\setminus\{0\}\to V$ specifies the parent of each non-root joint $j$ and satisfies $\mathsf{p}(j)<j$.
Each non-root joint $j$ identifies the bone connecting $\mathsf{p}(j)$ to $j$, yielding $J-1$ bones.
%
%
Each bone $j\in\{1,\ldots,J-1\}$ has a prescribed length $\ell_j>0$ that remains fixed during generation.
The prescribed lengths form the vector $\ell\triangleq(\ell_1,\ldots,\ell_{J-1}) \in \mathbb{R}^{J-1}$.
In our experiments, each prescribed bone length is the mean length of that bone over the observed frames.
The kinematic tree is the triple $\mathcal K\triangleq(V,\mathsf{p},\ell)$.

The observation and prediction windows contain $T_{\mathrm p}\in\mathbb N_+$ and $T_{\mathrm f}\in\mathbb N_+$ frames, respectively, representing discrete instances of time. 
The corresponding frame index sets are $\mathcal I_{\mathrm p}\triangleq\{-T_{\mathrm p}+1,\ldots,0\}$ and $\mathcal I_{\mathrm f}\triangleq\{1,\ldots,T_{\mathrm f}\}$.
The complete frame index set is defined as $\mathcal I\triangleq\mathcal I_{\mathrm p}\cup\mathcal I_{\mathrm f}$.
Consecutive frames are separated by a fixed period $\Delta>0$, measured in seconds.
Frame $t\in\mathcal I$ has physical time $t\Delta$, with time zero corresponding to the last observed frame.
For $t,q\in\mathcal I$, the signed time offset from frame $t$ to frame $q$ is $(q-t)\Delta$.
At frame $t$, the root state is a position $r_t\in\mathbb R^3$, and the non-root state $d_{t,j}\in\mathbb S^2$ is the direction of the bone from $\mathsf{p}(j)$ to $j$.
Thus, each node has a state space $\mathcal M_j$, with $\mathcal M_0 \triangleq\mathbb R^3$ and $\mathcal M_j \triangleq\mathbb S^2$ for $j=1,\ldots,J-1$.
Define the node states by $x_{t,0}\triangleq r_t$ and $x_{t,j}\triangleq d_{t,j}$ for $j\geq1$.
The Cartesian joint positions $p_{t,j}\in\mathbb R^3$ are reconstructed from the root position, bone directions, and bone lengths.
The forward kinematics of the skeleton are specified through the recursion
\begin{align}
p_{t,0}\triangleq r_t,\quad p_{t,j}\triangleq p_{t,\mathsf{p}(j)}+\ell_jd_{t,j},\quad j=1,\ldots,J-1.
\label{eq:forward_kinematics}
\end{align}

\begin{figure}[t]
\centering
\includegraphics[width=\linewidth]{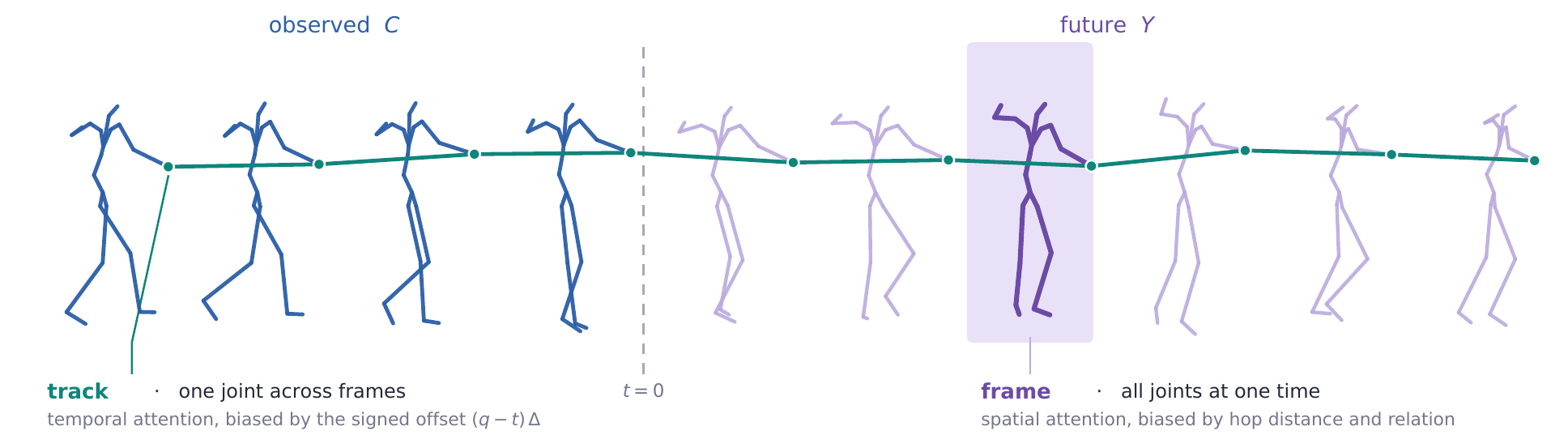}
\caption{Spatiotemporal skeleton graph with observed motion $C$ and recorded future motion $Y$.}
\label{fig:spatiotemporal_graph}
\end{figure}

The observed and future motion states of each skeleton form a spatiotemporal graph.
Each node of the graph is represented by a spatial coordinate $j$, representing the joint index, and a temporal coordinate, $t$, representing the frame index.
The node set is defined as $\mathcal V\triangleq\mathcal I\times V=\mathcal{V}_{\mathrm p}\cup\mathcal{V}_{\mathrm f}$, where the nodes from the observed trajectory are denoted by $\mathcal V_{\mathrm p}\triangleq\mathcal I_{\mathrm p}\times V$, and the nodes from the future trajectory are denoted by $\mathcal V_{\mathrm f}\triangleq\mathcal I_{\mathrm f}\times V$.
Each non-root joint is connected to its parent joint using spatial edges within the same frame.
We define the spatial edges as
\begin{align*}
\mathcal E_{\mathrm{sp}}\triangleq\big\{\{(t,\mathsf{p}(j)),(t,j)\}:t\in\mathcal I,\ j=1,\ldots,J-1\big\}.
\end{align*}
Temporal edges connect the same joint in consecutive frames.
We define the temporal edges as
\begin{align*}
\mathcal E_{\mathrm{tm}}\triangleq\big\{\{(t,j),(t+1,j)\}:t,t+1\in\mathcal I,\ j\in V\big\}.
\end{align*}
The resulting graph is $\mathcal G\triangleq(\mathcal V,\mathcal E_{\mathrm{sp}}\cup\mathcal E_{\mathrm{tm}})$.
We use the term `frame' to refer to all nodes with the same temporal index, and the term `joint track' to refer to all nodes having the same joint index.
The observed motion is represented by $C\triangleq(x_{t,j})_{(t,j)\in\mathcal V_{\mathrm p}}$.
The recorded future motion is represented by $Y\triangleq(x_{t,j})_{(t,j)\in\mathcal V_{\mathrm f}}$.
We illustrate the spatiotemporal graph structure in Figure~\ref{fig:spatiotemporal_graph} by highlighting a single representative frame and joint track.

A candidate future trajectory is represented by $X\triangleq(X_{t,j})_{(t,j)\in\mathcal V_{\mathrm f}}$, with $X_{t,j}\in\mathcal M_j$.
The state space and tangent space of the candidate future trajectory are given by the following products:
\begin{align*}
\mathcal X_{\mathcal G}\triangleq\prod_{(t,j)\in\mathcal V_{\mathrm f}}\mathcal M_j,\quad \mathcal{T}_X\mathcal X_{\mathcal G}\triangleq\prod_{(t,j)\in\mathcal V_{\mathrm f}} \mathcal{T}_{X_{t,j}}\mathcal M_j.
\end{align*}
Each element of $\mathcal X_{\mathcal G}$ contains the complete future trajectory.
%
%
Given the observed motion $C$ and kinematic tree $\mathcal K$, the task is to generate complete motion trajectories $X\in\mathcal X_{\mathcal G}$ over the next $T_{\mathrm f}\in \mathbb{N}_+$ frames.
%
%
%
Throughout this work, the root position is fixed at the origin in every observed and future frame, and each joint position is defined recursively with respect to the root.
%
%
The conditional distribution of the recorded future $Y$ is denoted by $p_Y(\cdot\mid C,\mathcal K)$.
The GSFM model induces a conditional distribution $p_X(\cdot\mid C,\mathcal K)$ over generated future trajectories.
As such, the learning objective is to match these two conditional distributions so that generated trajectories reproduce the variability and coordination of future motion conditioned on the observations.

\section{Graph Structured Flow Matching}
\label{sec:method}
In this section, we demonstrate how we generate a complete future motion trajectory by transporting all future joints and frames through a single conditional velocity field.
For a skeleton with $J\geq2$ joints, a neural network with parameters $\theta$ defines the velocity field $v_\theta$ on $\mathcal X_{\mathcal G}$, conditioned on the observed motion $C$ and kinematic tree $\mathcal K$.
For generation time $s\in[0,1]$, the evolving future trajectory is denoted by $\widehat X_s\in\mathcal X_{\mathcal G}$.
%
%
The conditional source distribution is denoted by $q_0(\cdot\mid C,\mathcal K)$ and the conditional flow satisfies the following differential equation:
\begin{align}
    \frac{d\widehat X_s}{ds}=v_\theta(\widehat X_s,s,C;\mathcal K)\in\mathcal T_{\widehat X_s}\mathcal X_{\mathcal G},\quad \widehat X_0\sim q_0(\cdot\mid C,\mathcal K).
    \label{eq:gsfm_dynamics}
\end{align}
Physical time $t\Delta$ indexes frames within $\widehat X_s$, whereas generation time $s$ parameterizes the evolution of the complete trajectory.
%
%
Equation~\ref{eq:gsfm_dynamics} specifies how bone directions evolve over generation time. We note that the root position is fixed at the origin for all observed and future frames.
The conditional flow evaluated at $s=1$, denoted by $\widehat{X}_1$, is the generated future trajectory. 
$\widehat{X}_1$ has the conditional distribution $p_X(\cdot\mid C,\mathcal K)$.
We train the parameterized velocity field $v_\theta$ by conditional flow matching along geodesic paths that connect source trajectories to recorded future trajectories.
%

Section~\ref{sec:conditional_training} defines the source distribution, constructs geodesic paths connecting source trajectories to recorded futures, and formulates the velocity regression objective.
Section~\ref{sec:velocity_field} constructs the graph structured velocity field $v_\theta$ and explains how it is (i) trained by velocity regression and (ii) integrated to generate future trajectories.

\subsection{Conditional Flow Matching}
\label{sec:conditional_training}
For each bone $j\in V\setminus\{0\}$, $C_{0,j}$ denotes its direction in the final observed frame (i.e., at $t=0$).
To sample a source trajectory, we repeat the final observed pose across all future frames and independently perturb each repeated bone direction.
For every $t\in\mathcal I_f$ and $j\in V\setminus\{0\}$, sample $\xi_{t,j}\sim\mathcal N(0,I_3)$ independently across frames and bones, and independently of $Y$ conditional on $(C,\mathcal K)$.
The source trajectory $Z$ uses a fixed tangent perturbation scale $\sigma_{\mathrm d}>0$ with
\begin{align}
Z_{t,0}\triangleq0,\quad Z_{t,j}\triangleq\mathbf{exp}_{C_{0,j}}\!\left(\sigma_{\mathrm d}\Pi_{C_{0,j}}(\xi_{t,j})\right),\quad j\in V\setminus\{0\}.
\label{eq:source}
\end{align}
Throughout the rest of the manuscript, we refer to the parameter $\sigma_d$ as the `source scale.' The conditional distribution of $Z$ given $(C,\mathcal K)$ defines the source distribution $q_0(\cdot\mid C,\mathcal K)$.
%
%
The sampled trajectory $Z$ provides the initial condition $\widehat X_0=Z$ for the flow in Equation~\ref{eq:gsfm_dynamics}.
%
%
Source trajectories are sampled using the same value of $\sigma_{\mathrm d}$ during training and generation.
%
%
The source trajectory $Z$ and recorded future $Y$ define a geodesic path $X_s$ in the product space $\mathcal X_{\mathcal G}$.
The path derivative $U_s^\star$ provides the target velocity that we use in the subsequent regression task:
\begin{align}
    X_s\triangleq\mathbf{exp}_Z\!\left(s\mathbf{log}_Z(Y)\right),\quad U_s^\star\triangleq\frac{dX_s}{ds}=\mathcal T_{Z\to X_s}\!\left(\mathbf{log}_Z(Y)\right).
    \label{eq:conditional_path}
\end{align}
The training path $X_s$ is computed directly from $Z$ and $Y$, whereas $\widehat X_s$ follows the learned dynamics in Equation~\ref{eq:gsfm_dynamics}.
%
%
%
%
The interpolation of $X_s$ satisfies $X_0=Z$ and $X_1=Y$, and its velocity is tangent at $X_s$.
For the spherical logarithm map, we require $Z_{t,j}\neq-Y_{t,j}$ since antipodal directions do not determine a unique minimizing geodesic.

Consider a candidate velocity field $v$ that is equal to zero at the root joint and satisfies $v(X,s,C;\mathcal K)\in\mathcal T_X\mathcal X_{\mathcal G}$.
Assume that $\mathbb E[\|v(X_s,s,C;\mathcal K)\|^2]<\infty$ under the sampling scheme above.
The ideal velocity regression objective is defined as 
\begin{align}\label{eq:ideal_fm_loss}
    \mathcal L_{\mathrm{FM}}(v)\triangleq\frac{1}{(J-1)T_{\mathrm f}}\mathbb E\!\left[\left\|v(X_s,s,C;\mathcal K)-U_s^\star\right\|^2\right].
\end{align}
%
%
%
The conditional mean field minimizes $\mathcal L_{\mathrm{FM}}$ and is defined as follows
\begin{align}\label{eq:conditional_mean}
    v^\star(X,s,C;\mathcal K)\triangleq\mathbb E\!\left[U_s^\star\mid X_s=X,s,C,\mathcal K\right].
\end{align}
Each component of $v^\star$ conditions on all future joints and frames in $X$, rather than only on the corresponding bone direction.
The resulting field can therefore represent dependencies across joints and frames.
%
%
Appendix~\ref{app:conditional_analysis} proves the regression identity and states the existence and uniqueness assumptions for conditional marginal transport.
Under these assumptions, the flow induced by $v^\star$ has the same conditional marginals as $X_s$ and terminal distribution $p_Y(\cdot\mid C,\mathcal K)$.
%

\subsection{Graph Structured Velocity Field}
\label{sec:velocity_field}
We construct the velocity field $v_\theta$ to couple the evolution of future bone directions across joints and physical frames.
For each input $(X,s,C,\mathcal K)$, the neural network jointly processes the observed motion $C$ and candidate future skeletal trajectory $X$ on the spatiotemporal graph $\mathcal G$.
Spatial attention combines features from all joints within a frame, using learned biases based on skeletal hop distance and joint relations.
Temporal attention then combines the spatially updated features across all observed and future frames of each joint track, using a learned bias based upon the signed physical time offset between frames.
%
%
%
%
%
We then compute the velocity field $v_\theta (X,s,C;\mathcal{K})$ using the spatial and temporal attention mechanism.
%
%
This structure enables the spatiotemporal graph structure to influence the evolution of the future trajectory during every generation step.

\begin{algorithm}[t]
\caption{One GSFM training step. Every operation acts elementwise over examples, future frames $t\in\mathcal I_f$, and bones $j\in V\setminus\{0\}$.}
\label{alg:gsfm_training}
\begin{algorithmic}[1]
\Require batch $\{(C_i,Y_i,\mathcal K_i)\}_{i=1}^{m}$ with $m\in\mathbb N_+$, parameters $\theta$, source scale $\sigma_d$, exclusion threshold $\varepsilon_{\mathrm{cut}}$
\State $\xi_{t,j}\sim\mathcal N(0,I_3)$;\quad $Z_{t,0}\gets 0$,\; $Z_{t,j}\gets \exp_{C_{0,j}}\!\big(\sigma_d\,\Pi_{C_{0,j}}(\xi_{t,j})\big)$ \Comment{source, Equation~\ref{eq:source}}
\State $\zeta\sim\mathcal N(0,1)$;\quad $s\gets (1+e^{-\zeta})^{-1}$
\State $X_s\gets \exp_Z\!\big(s\log_Z Y\big)$,\quad
       $U\gets \mathcal T_{Z\to X_s}\!\big(\log_Z Y\big)$ \Comment{path and target, Equation~\ref{eq:conditional_path}}
\State $\mathcal A\gets\{(i,t,j):i\in\{1,\ldots,m\},\ t\in\mathcal I_f,\ j\in V\setminus\{0\},\ \langle Z_{i,t,j},Y_{i,t,j}\rangle\geq-1+\varepsilon_{\mathrm{cut}}\}$
\State $\widehat{\mathcal L}_{\mathrm{FM}}(\theta)\gets\displaystyle\frac{\sum_{(i,t,j)\in\mathcal A}\|v_\theta(Q_i,s_i,C_i;\mathcal K_i)_{t,j}-U_{i,t,j}\|^2}{\max\{|\mathcal A|,1\}}$
\State Update $\theta$ using $\nabla_\theta\widehat{\mathcal L}_{\mathrm{FM}}(\theta)$.
\end{algorithmic}
\end{algorithm}


\begin{algorithm}[t]
\caption{GSFM generation.}
\label{alg:gsfm_generation}
\begin{algorithmic}[1]
\Require observed motion $C$, kinematic tree $\mathcal K$, trained parameters $\theta$, source scale $\sigma_d$, number of steps $N$
\State draw $X\sim q_0(\cdot\mid C,\mathcal K)$ as in Algorithm~\ref{alg:gsfm_training}, line~1;\quad $\eta\gets 1/N$
\For{$n=0,\dots,N-1$}
  \State $s\gets n\eta$;\quad $V\gets v_\theta(X,s,C;\mathcal K)$;\quad $X_{\rm mid}\gets \exp_X\!\big(\tfrac{\eta}{2}V\big)$
  \State $V\gets \Pi_X\!\Big(\mathcal T_{X_{\rm mid}\to X}\, v_\theta\big(X_{\rm mid},\,s+\tfrac{\eta}{2},\,C;\mathcal K\big)\Big)$
  \State $X\gets \mathcal R\big(\exp_X(\eta V)\big)$
\EndFor
\State \Return joint positions reconstructed from $X$ by Equation~\ref{eq:forward_kinematics}
\end{algorithmic}
\end{algorithm}


Recall that the vector $\theta$ denotes all tunable weights in the GSFM model.
We denote the network output before projection by $a_\theta(X,s,C;\mathcal K)_{t,j}\in\mathbb R^3$ for each future node $(t,j)\in\mathcal{V}_{\mathrm f}$.
%
%
For a future bone direction $X_{t,j}\in\mathbb S^2$, a valid velocity lies in $T_{X_{t,j}}\mathbb S^2$.
For every $t\in\mathcal I_{\mathrm f}$, we therefore define the velocity field $v_\theta$ by setting the root output to zero and projecting each bone output onto the associated tangent space:
\begin{align}
v_\theta(X,s,C;\mathcal K)_{t,0}
\triangleq0,\quad v_\theta(X,s,C;\mathcal K)_{t,j}
\triangleq\Pi_{X_{t,j}}\!\left(a_\theta(X,s,C;\mathcal K)_{t,j}\right),
\quad j\in V\setminus\{0\}.
\label{eq:tangent_readout}
\end{align}
In Appendix~\ref{app:network_architecture}, we specify the node features and neural network operations which produce $a_\theta(X,s,C;\mathcal K)_{t,j}$.

The ideal objective for $\theta$ is quantified by the loss function $\mathcal L_{\mathrm{FM}}(v_\theta)$, which is obtained by setting $v=v_\theta$ in Equation~\ref{eq:ideal_fm_loss}.
At each sampled training input $(X_s,s,C,\mathcal K)$, this objective compares the predicted tangent vector $v_\theta(X_s,s,C;\mathcal K)$ with the target tangent vector $U_s^\star$.
The loss is evaluated after applying the projection in Equation~\ref{eq:tangent_readout}, and the gradient of the loss with respect to the parameters is used to update $\theta$.
In implementation, Algorithm~\ref{alg:gsfm_training} updates $\theta$ using a numerical batch loss $\widehat{\mathcal L}_{\mathrm{FM}}(\theta)$.
We address numerical geometric operations, the construction of numerical paths, exclusion of endpoint pairs near antipodal configurations, and optimization settings in Appendix~\ref{app:numerical_implementation}.
%
%
%
%
%

\section{Experiments}
\label{sec:experiments}

\subsection{Experimental Setup}
\label{sec:experimental_setup}
We trained the GSFM model on the AMASS dataset~\citep{mahmood2019amass} in which each supplied skeleton is composed of $22$ joints.
%
%
To assess the generalizability of GSFM, we study zero-shot transfer to the Human3.6M~\citep{ionescu2014human36m} dataset, in which each supplied skeleton is composed of $17$ joints.
The root position is fixed at the origin for the duration of each generated trajectory.
Each observed trajectory contains $30$ frames, and we generate future trajectories over a horizon of $120$ frames, representing $2$ seconds of physical time.
For each test observation, we generate 50 future trajectories and evaluate them using the metrics in Appendix~\ref{app:experimental_protocols}.

GSFM-deep processes node features through $12$ sequential blocks each consisting of a spatial attention layer, temporal attention layer, and a feedforward layer during evaluation of $v_\theta$.
Figure~\ref{fig:velocity_block} illustrates the structure of one block.
Each block has distinct attention and feedforward parameters.
GSFM-tied instead reuses one block $12$ times with shared attention and feedforward parameters.
Unless stated otherwise, GSFM uses a source scale of $\sigma_{\mathrm d}=0.7$ during training and generation, with $25$ midpoint steps during generation.
NFE denotes the number of velocity field evaluations, giving $50$ NFE for the primary configuration.
Appendices~\ref{app:network_architecture} and~\ref{app:numerical_implementation} specify the architecture and optimization settings.
We adopt the evaluation metrics used in EquiFusion~\citep{curreli2026equifusion}.
Appendix~\ref{app:experimental_protocols} specifies the data protocols, metric definitions and reductions, and comparison settings.
In all subsequent tables, the designation (A) denotes that a model was trained on the AMASS dataset, and ``\textemdash'' denotes an unreported or inapplicable entry.
Bold metric values identify the lowest reported errors within each comparison, including ties at the displayed precision.

\begin{table}[t]
\centering
\caption{Prediction accuracy, diversity calibration, motion statistics, and bone length consistency on the AMASS dataset.
Baseline results are reproduced from EquiFusion~\citep{curreli2026equifusion}.
Str. and Jit. are reported as percentages, and $0.00$ denotes zero at the displayed precision.
Bold metric values denote the best performance in this table and following all tables.}
\label{tab:amass_results}
\setlength{\tabcolsep}{3pt}
\scriptsize
\resizebox{\linewidth}{!}{%
\begin{tabular}{lcccccccccc}
\toprule
& \multicolumn{3}{c}{Prediction}
& \multicolumn{2}{c}{Multimodal prediction}
& \multicolumn{2}{c}{Diversity}
& \multicolumn{1}{c}{\shortstack{Motion statistics}}
& \multicolumn{2}{c}{\shortstack{Bone length\\consistency}} \\
\cmidrule(lr){2-4}
\cmidrule(lr){5-6}
\cmidrule(lr){7-8}
\cmidrule(lr){9-9}
\cmidrule(lr){10-11}
Method
& ADE $\downarrow$
& FDE $\downarrow$
& MAE $\downarrow$
& MMADE $\downarrow$
& MMFDE $\downarrow$
& APDE $\downarrow$
& APD
& CMD $\downarrow$
& Str. $\downarrow$
& Jit. $\downarrow$ \\
\midrule
ZeroVelocity & 0.755 & 0.992 & 7.779 & 0.814 & 1.015 & \textemdash & 0.000 & 39.262 & \textbf{0.00} & \textbf{0.00} \\
TPK & 0.656 & 0.675 & 10.191 & 0.658 & 0.674 & 2.265 & 9.283 & 17.127 & 7.34 & 0.34 \\
DLow & 0.590 & 0.612 & 8.510 & 0.618 & 0.617 & 4.243 & 13.170 & 15.185 & 8.41 & 0.40 \\
GSPS & 0.563 & 0.613 & 9.045 & 0.609 & 0.633 & 4.678 & 12.465 & 18.404 & 6.65 & 0.29 \\
DivSamp & 0.564 & 0.647 & 8.027 & 0.623 & 0.667 & 15.837 & 24.724 & 50.239 & 11.17 & 0.82 \\
BeLFusion & 0.513 & 0.560 & 7.125 & 0.569 & 0.585 & \textbf{1.977} & 9.376 & 16.995 & 7.19 & 0.34 \\
CoMusion & 0.494 & 0.547 & 6.715 & \textbf{0.469} & \textbf{0.466} & 2.328 & 10.848 & 9.636 & 4.04 & 0.25 \\
SkeletonDiffusion & 0.480 & \textbf{0.545} & 6.124 & 0.561 & 0.580 & 2.067 & 9.456 & 11.417 & 3.15 & 0.20 \\
EquiFusion (A) & 0.496 & 0.560 & 6.214 & 0.576 & 0.596 & 2.518 & 8.241 & 13.097 & \textbf{0.00} & \textbf{0.00} \\
\midrule
\textbf{GSFM-deep} & \textbf{0.477} & 0.550 & \textbf{5.840} & 0.578 & 0.597 & 2.840 & 8.244 & 5.614 & \textbf{0.00} & \textbf{0.00} \\
\textbf{GSFM-tied} & 0.484 & 0.551 & 5.930 & 0.583 & 0.598 & 2.683 & 8.528 & \textbf{4.776} & \textbf{0.00} & \textbf{0.00} \\
\bottomrule
\end{tabular}}
\end{table}

\subsection{Performance on AMASS and Zero-Shot Transfer to Human3.6M}
\label{sec:amass_results}
Table~\ref{tab:amass_results} compares prediction accuracy, diversity calibration, motion statistics, and bone length consistency on AMASS.
GSFM-deep attains the lowest ADE and MAE, while GSFM-tied achieves the lowest CMD among the compared methods. Their multimodal errors are comparable to EquiFusion, although CoMusion remains stronger on these metrics and GSFM-deep ranks third in FDE.
These results show that competitive prediction and agreement with ground truth trajectories can be achieved while preserving prescribed bone lengths during generation.
The left panel of Figure~\ref{fig:teaser} illustrates different future trajectories generated from the same observed motion.
The right panel shows that despite having substantially fewer parameters, GSFM-tied has comparable AMASS ADE performance, while Table~\ref{tab:amass_results} reports the corresponding prediction and multimodal errors.

Next, we evaluate the ability of GSFM to generalize to another dataset having a different skeleton structure.
Both GSFM configurations (-deep and -tied) are trained on AMASS and then applied directly to the Human3.6M skeleton without parameter updates or retargeting.
To obtain the results in Table~\ref{tab:zero_shot}, bone lengths and structural node features were obtained using the kinematic tree for the Human3.6M dataset.
%
%
For both GSFM configurations, the learned parameters $\theta$ and source scale $\sigma_d$ remain unchanged.
%
%
Table~\ref{tab:zero_shot} shows that both GSFM configurations (-deep and -tied) achieve lower ADE and FDE than the every baseline that uses retargeting.
However, we note that EquiFusion outperforms GSFM in displacement error statistics.
In a similar fashion to the results in Table~\ref{tab:amass_results}, GSFM-tied obtains a comparable displacement accuracy to GSFM-deep, while having only $4.47$ million parameters.
%

\begin{table}[t]
\centering
\caption{Zero-shot transfer from the AMASS dataset to the Human3.6M dataset.
%
%
The label ``retarget'' identifies learning pipelines which convert between representations for the training and target skeletons.
%
}
\label{tab:zero_shot}
\setlength{\tabcolsep}{3pt}
\scriptsize
\resizebox{\linewidth}{!}{%
\begin{tabular}{lcccccccc}
\toprule
& \multicolumn{3}{c}{Prediction}
& \multicolumn{2}{c}{Multimodal prediction}
& \multicolumn{2}{c}{Diversity}
& \multicolumn{1}{c}{\shortstack{Motion statistics}} \\
\cmidrule(lr){2-4}
\cmidrule(lr){5-6}
\cmidrule(lr){7-8}
\cmidrule(lr){9-9}
Method
& ADE $\downarrow$
& FDE $\downarrow$
& MAE $\downarrow$
& MMADE $\downarrow$
& MMFDE $\downarrow$
& APDE $\downarrow$
& APD
& CMD $\downarrow$ \\
\midrule
ZeroVelocity & 0.597 & 0.884 & 6.753 & 0.683 & 0.909 & 8.085 & 0.000 & 22.812 \\
TPK + retarget & 1.154 & 0.983 & 22.686 & 1.155 & 0.987 & \textbf{1.968} & 7.221 & 10.051 \\
DLow + retarget & 1.094 & 0.948 & 22.272 & 1.096 & 0.951 & 2.060 & 9.683 & 9.204 \\
GSPS + retarget & 1.193 & 1.051 & 11.439 & 1.194 & 1.053 & 2.373 & 9.985 & 7.409 \\
BeLFusion + retarget & 1.226 & 1.029 & 10.957 & 1.225 & 1.033 & 2.284 & 6.483 & 8.031 \\
CoMusion + retarget & 1.221 & 1.028 & 22.521 & 1.218 & 1.032 & 2.370 & 9.926 & 8.587 \\
SkeletonDiffusion + retarget & 1.372 & 1.193 & 17.232 & 1.371 & 1.194 & 2.995 & 5.420 & 7.616 \\
EquiFusion (A) & \textbf{0.403} & \textbf{0.533} & \textbf{5.861} & \textbf{0.536} & \textbf{0.585} & 2.248 & 9.320 & 7.061 \\
\midrule
\textbf{GSFM-deep (A)} & 0.455 & 0.645 & 6.162 & 0.596 & 0.696 & 2.895 & 9.452 & \textbf{5.060} \\
\textbf{GSFM-tied (A)} & 0.484 & 0.660 & 6.334 & 0.612 & 0.709 & 4.364 & 12.102 & 11.607 \\
\bottomrule
\end{tabular}}
\end{table}

\subsection{Ablation Studies}
\label{sec:graph_ablations}

\textbf{Spatial and temporal interactions.}
Table~\ref{tab:graph_ablations} evaluates how the inclusion of spatial and temporal information affects the quality of the predicted velocity field $v_\theta$.
Each restriction in the ablation study changes only the corresponding attention mask for one-hop restrictions.
The full configuration uses spatial attention over all joints within each frame and temporal attention over all observed and future frames of the same joint.
%
%
The spatial one-hop configuration restricts spatial attention to the current joint and its immediate skeletal neighbors, while temporal attention remains full.
The temporal one-hop configuration restricts temporal attention to the current and adjacent frames, while spatial attention remains full.
%

Both spatial and temporal one hop restrictions increase ADE and FDE relative to full attention.
These comparisons support direct information exchange beyond immediate skeletal neighbors and adjacent frames within each block.
Removing either attention sublayer increases ADE further than its corresponding one hop restriction.
In particular, removing temporal attention achieves APD comparable to the full attention configuration, but at the cost of substantially increased CMD.
%
%
Together, these results support combining information across joints and frames when constructing the velocity field $v_\theta$ for the complete future trajectory.

\begin{table}[t]
\centering
\caption{Contribution of spatial and temporal attention to trajectory accuracy and motion statistics on AMASS.
%
%
%
Let $\Delta$ADE denote the percentage increase in ADE relative to the `full spatial and temporal attention' configuration.}
\label{tab:graph_ablations}
\setlength{\tabcolsep}{4pt}
\small
\resizebox{\linewidth}{!}{%
\begin{tabular}{lccccc}
\toprule
& \multicolumn{2}{c}{Prediction}
& \multicolumn{1}{c}{Diversity}
& \multicolumn{1}{c}{\shortstack{Motion statistics}}
& \\
\cmidrule(lr){2-3}
\cmidrule(lr){4-4}
\cmidrule(lr){5-5}
Variant
& ADE $\downarrow$
& FDE $\downarrow$
& APD
& CMD $\downarrow$
& $\Delta$ADE $\downarrow$ \\
\midrule
Full spatial and temporal attention & \textbf{9.474} & \textbf{10.33} & 18.4 & \textbf{8.6} & \textemdash \\
Spatial one hop & 10.129 & 11.35 & 18.9 & 15.0 & $6.9\%$ \\
Spatial attention removed & 10.824 & 12.65 & 17.2 & 14.6 & $14.3\%$ \\
Temporal one hop & 12.073 & 12.08 & 12.7 & 32.4 & $27.4\%$ \\
Temporal attention removed & 14.441 & 11.27 & 18.4 & 482.2 & $52.4\%$ \\
\bottomrule
\end{tabular}}
\end{table}

\textbf{Parameter sharing.}
Table~\ref{tab:parameter_sharing} compares GSFM-deep and GSFM-tied using the same source scale and computational depth.
Their block architectures are identical, but the attention and feedforward parameters are either distinct or shared across each block.
Sharing the attention and feedforward parameters substantially reduces the parameter count ($30.49\mathrm{M}\to 4.47\mathrm{M}$), while maintaining comparable displacement accuracy to GSFM-deep.
It is evident from Table~\ref{tab:parameter_sharing} that distinct attention and feedforward parameters at every block application may not be required to enable competitive trajectory prediction.
%
%
Figure~\ref{fig:teaser} compares ADE and parameter count for the baselines with reported parameter counts.

\textbf{Source scale.}
Each row in Table~\ref{tab:parameter_sharing} corresponds to a separately trained model that uses the same source scale during training and generation.
In our testing, we did not find a single value of $\sigma_d$ which achieved the best performance across all tested metrics.
Intermediate values ($\sigma_d = 0.5, 0.7$) lower ADE by $1.3 - 1.6\%$ relative to $\sigma_d = 0.9$ without degrading MMADE or MMFDE, at the cost of lower sample diversity and higher APDE (GSFM-deep: APD $8.244$ vs $8.585$, APDE $2.840$ vs $2.513$).
The observed results reflect a trade-off between prediction error and distributional statistics, rather than a uniform improvement from increasing or decreasing the value of the source scale.
%
%
Appendix~\ref{app:inference_cost} reports the sampling time and memory of GSFM-deep at 50 and 20 NFE, together with the resulting change in prediction metrics.

\begin{table}[t]
\centering
\caption{In-depth comparison between GSFM-deep and GSFM-tied over varying source scale.
For each model, the source scale value is used during both training and generation. 
Parameter counts are in millions, denoted by $M$.}
\label{tab:parameter_sharing}
\setlength{\tabcolsep}{3pt}
\scriptsize
\resizebox{\linewidth}{!}{%
\begin{tabular}{lcccccccccc}
\toprule
\multicolumn{3}{c}{Configuration}
& \multicolumn{3}{c}{Prediction}
& \multicolumn{2}{c}{Multimodal prediction}
& \multicolumn{2}{c}{Diversity}
& \multicolumn{1}{c}{\shortstack{Motion statistics}} \\
\cmidrule(lr){1-3}
\cmidrule(lr){4-6}
\cmidrule(lr){7-8}
\cmidrule(lr){9-10}
\cmidrule(lr){11-11}
Model
& Parameters (M)
& $\sigma_{\mathrm d}$
& ADE $\downarrow$
& FDE $\downarrow$
& MAE $\downarrow$
& MMADE $\downarrow$
& MMFDE $\downarrow$
& APDE $\downarrow$
& APD
& CMD $\downarrow$ \\
\midrule
Tied & 4.47 & 0.5 & 0.484 & 0.550 & 5.917 & 0.582 & \textbf{0.595} & 2.638 & 8.554 & 5.246 \\
Tied & 4.47 & 0.7 & 0.484 & 0.551 & 5.930 & 0.583 & 0.598 & 2.683 & 8.528 & \textbf{4.776} \\
Tied & 4.47 & 0.9 & 0.492 & 0.558 & 6.020 & 0.588 & 0.601 & \textbf{2.213} & 8.890 & 4.903 \\
\midrule
Deep & 30.49 & 0.3 & 0.485 & 0.553 & 5.856 & 0.582 & 0.597 & 2.822 & 8.263 & 6.521 \\
Deep & 30.49 & 0.5 & \textbf{0.477} & \textbf{0.549} & \textbf{5.815} & \textbf{0.577} & 0.596 & 2.876 & 8.187 & 5.943 \\
Deep & 30.49 & 0.7 & \textbf{0.477} & 0.550 & 5.840 & 0.578 & 0.597 & 2.840 & 8.244 & 5.614 \\
Deep & 30.49 & 0.9 & 0.484 & 0.552 & 5.910 & 0.583 & 0.597 & 2.513 & 8.585 & 4.964 \\
\bottomrule
\end{tabular}}
\end{table}

\section{Conclusion}
In this work, we introduced a \emph{Graph Structured Flow Matching} (GSFM) model for human motion prediction.
The developed method transports the complete future skeletal trajectory through a single conditional velocity field while accounting for spatial and temporal interactions.
A representation of the skeleton based upon unit bone directions preserves input bone lengths during trajectory generation.
Experiments on the AMASS dataset demonstrate the contributions of spatial and temporal communication to prediction accuracy and motion statistics.
Sharing attention and feedforward parameters across the 12 blocks retains model depth while substantially reducing the parameter count.
Models trained on AMASS are shown to transfer directly to the Human3.6M skeleton without retargeting or parameter updates, demonstrating transferrability to unseen data and skeletal structures.
%
%
Future work will include improving transfer learning performance and reducing the computational cost of trajectory generation.

\newpage
\subsection*{AI use statement}
In this work, we have not used generative AI tools for any tasks with required disclosure.
We used generative AI tools solely for language editing to improve the grammar, clarity, and readability of the manuscript.
We have reviewed all work assisted by generative AI.
Specifically, we checked all suggested language edits to ensure that they preserved the intended meaning and technical accuracy.
We take responsibility for the final content of this work, including text, claims or artifacts produced with the aid of generative AI.

\bibliography{ref}
\bibliographystyle{plainnat}

\newpage
\appendix

\section{Conditional Flow Matching Analysis}
\label{app:conditional_analysis}
%
%

\subsection{Source Distribution and Path Domain}\label{subapp:source_dist}
The symbol $\mathbb{P}$ denotes probability under the data and source sampling distributions.
Fix the conditioning pair $(C,\mathcal K)$ and a node $(t,j)\in\mathcal I_{\mathrm f}\times(V\setminus\{0\})$.
For $p\triangleq C_{0,j}$ and $\xi\sim\mathcal N(0,I_3)$, the source perturbation is $W\triangleq\sigma_{\mathrm d}\Pi_p(\xi)\in\mathcal T_p\mathbb S^2$.
Its coordinates in any orthonormal tangent basis are independent Gaussian variables with variance $\sigma_{\mathrm d}^2$.
In these coordinates, the density $f_W$ has the following expression:
\begin{align*}
    f_W(w)\triangleq\frac{1}{2\pi\sigma_{\mathrm d}^2}\exp\!\left(-\frac{\|w\|^2}{2\sigma_{\mathrm d}^2}\right),\quad w\in\mathbb R^2.
\end{align*}
For any fixed radius, the corresponding circle has zero planar measure, so $\|W\|$ assigns probability zero to each fixed value and to every countable set.

For any $q\in\mathbb S^2$, the event $\mathbf{exp}_p(W)=q$ requires that $\cos\|W\|=\langle p,q\rangle$.
The set of radii $r=\lVert W \rVert$ which satisfy $\cos\|W\|=\langle p,q\rangle$ form a countable set.
As a result, 
\begin{align*}
    \mathbb P\!\left(\mathbf{exp}_p(W)=q\mid C,\mathcal K\right)\leq\mathbb P\!\left(\cos\|W\|=\langle p,q\rangle\mid C,\mathcal K\right)=0.
\end{align*}
Further, conditional independence of $Z$ and $Y$ implies that
\begin{align*}
    \mathbb P\!\left(Z_{t,j}=-Y_{t,j}\mid C,\mathcal K\right)
    &=\mathbb E\!\left[\mathbb P\!\left(Z_{t,j}=-Y_{t,j}\mid Y,C,\mathcal K\right)\mid C,\mathcal K\right] \\
    &=0.
\end{align*}
Since there are finitely many future bone nodes, no endpoint pair is antipodal with probability one.

For every such pair, define the geodesic angle $\vartheta_{t,j}\triangleq\arccos\langle Z_{t,j},Y_{t,j}\rangle\in[0,\pi)$.
If the endpoints coincide, their logarithm is zero and their path is constant.
Otherwise, the corresponding minimizing geodesic is unique.
Thus, Equation~\ref{eq:conditional_path} defines the complete path for every $s\in[0,1]$ almost surely.
The norm of the velocity of each bone node is preserved under parallel transport.
That is,
\begin{align*}
    \|U_{s,t,j}^\star\|=\left\|\mathbf{log}_{Z_{t,j}}(Y_{t,j})\right\|=\vartheta_{t,j}<\pi.
\end{align*}
Since the root nodes are fixed at the origin for all generation time $s\in[0,1]$, we have
\begin{align}\label{eq:u-star-bound}
    \lVert U_s^\star\rVert^2=\sum_{t\in\mathcal I_{\mathrm f}}\sum_{j\in V\setminus\{0\}}\vartheta_{t,j}^2<(J-1)T_{\mathrm f}\pi^2,
\end{align}
and the velocity has a finite second moment. 
%

\subsection{Conditional Mean and Regression Identity}\label{subapp:conditional_mean}
All expectations in this subsection follow the sampling distribution used in Equation~\ref{eq:ideal_fm_loss}.
Recall that the velocity field $v^\star$ is defined according to the conditional expectation in Equation~\ref{eq:conditional_mean}.
Conditioning on $X_s=x, C, \mathcal{K}$ fixes the trajectory and its tangent space.
For every node $(t,j)$, its associated conditional mean is orthogonal to its bone direction vector.
That is,
\begin{align}\label{eq:apply-jensen-here}
	\left\langle X_{t,j},v^\star(X,s,C;\mathcal K)_{t,j}\right\rangle =\mathbb E\!\left[\langle X_{s,t,j},U_{s,t,j}^\star\rangle\mid X_s=X,s,C;\mathcal K\right]=0.
\end{align}
The root nodes are fixed at the origin, and thus have zero conditional mean.
Applying conditional Jensen's inequality to Equation~\ref{eq:conditional_mean} and using Equation~\ref{eq:u-star-bound} yields
\begin{align*}
	\mathbb E\!\left[\|v^\star(X_s,s,C,\mathcal K)\|^2\right]\leq\mathbb E\!\left[\|U_s^\star\|^2\right]<\infty,
\end{align*}
implying that $v^\star$ is square integrable under the regression sampling distribution, and is thus admissible.

Our objective is to quantify $\mathbb{E}[\lVert v(X_s,s,C,\mathcal{K}) - U_s^\star \rVert^2]$, where $v(X_s,s,C,\mathcal{K})$ is any other square integrable tangent field.
Let $\Delta_v$ denote the difference between $v^\star$ and $v$ and define the error between the target field velocity and $v^\star$ as
\begin{align*}
	\Delta_v\triangleq v(X_s,s,C,\mathcal{K})-v^\star (X_s,s,C,\mathcal{K}),\quad  E^\star\triangleq U_s^\star - v^\star(X_s,s,C,\mathcal{K}).
\end{align*}
Adding and subtracting $v^\star (X_s,s,C,\mathcal{K})$ yields
\begin{align}
	\mathbb{E}[\lVert v(X_s,s,C,\mathcal{K}) - U_s^\star \rVert^2]&=\mathbb{E} [\lVert \Delta_v - E^\star \rVert^2] \nonumber \\
	&= \mathbb{E} [\lVert \Delta_v \rVert^2] + \mathbb{E} [\lVert E^\star \rVert^2] -2\mathbb{E} [\langle \Delta_v, E^\star \rangle] \label{eq:apply-here}.
\end{align}
Since $\Delta_v$ is a measurable function of $(X_s,s,C,\mathcal K)$ and both $\Delta_v$ and $E^\star$ are square integrable, $\langle\Delta_v,E^\star\rangle$ is integrable.
The definition of $v^\star$ in Equation~\ref{eq:conditional_mean} gives
\begin{align*}
\mathbb E[E^\star\mid X_s,s,C,\mathcal K]
&=\mathbb E[U_s^\star\mid X_s,s,C,\mathcal K]
-v^\star(X_s,s,C;\mathcal K)=0
\end{align*}
almost surely.
Taking the conditional expectation therefore yields
%
%
%
\begin{align}\label{eq:equals-zero}
	\mathbb{E} [\langle \Delta_v, E^\star \rangle]=\mathbb{E} [\langle \Delta_v,\mathbb{E} [E^\star \mid X_s,s,C,\mathcal{K}] \rangle]=0.
\end{align}
Applying Equation~\ref{eq:equals-zero} and the facts that $\mathbb{E} [\lVert v(X_s,s,C,\mathcal{K}) - U_s^\star \rVert^2]=(J-1)T_{\mathrm{f}} \mathcal{L}_{\mathrm{FM}} (v)$ and $\mathbb{E} [\lVert v^\star(X_s,s,C,\mathcal{K}) - U_s^\star \rVert^2]=(J-1)T_{\mathrm{f}} \mathcal{L}_{\mathrm{FM}} (v^\star)$ to Equation~\ref{eq:apply-here} gives
\begin{align*}
    \mathcal{L}_{\mathrm{FM}}(v)-\mathcal{L}_{\mathrm{FM}}(v^\star)=\frac{\mathbb{E}[\lVert v(X_s,s,C,\mathcal{K}) - v^\star(X_s,s,C,\mathcal{K}) \rVert^2]}{(J-1)T_{\mathrm{f}}}\geq 0.
\end{align*}
%
Hence, $\mathcal L_{\mathrm{FM}}(v)\geq\mathcal L_{\mathrm{FM}}(v^\star)$, with equality if and only if $v(X_s,s,C;\mathcal K)=v^\star(X_s,s,C;\mathcal K)$ almost surely under the regression sampling distribution.

\subsection{Conditional Marginal Transport}
Fix $(C,\mathcal{K})$ and let the results in Appendices~\ref{subapp:source_dist} and \ref{subapp:conditional_mean} hold. The resultant trajectories are defined on the following compact manifold:
\begin{align*}
    \mathcal{X}_{\mathcal{G}}^0\triangleq \{ X\in\mathcal{X}_{\mathcal{G}} : X_{t,0}=0\text{ for every }t\in\mathcal{I}_{\mathrm{f}} \},
\end{align*}
where $X_{t,0}$ denotes the root position at physical time $t$.
Denote the conditional distribution of the path $X_s$ on $\mathcal{X}_{\mathcal{G}}^0$ by $\rho_s$.
Then, consider a smooth real function $\varphi$ on $\mathcal{X}_{\mathcal{G}}^0$ having gradient $\nabla\varphi$ under the product metric.
We have that $\nabla\varphi$ is bounded due to the fact that $\mathcal{X}_{\mathcal{G}}^0$ is compact.

Applying the chain rule yields
\begin{align}\label{eq:chain-rule-deriv}
    \frac{d}{ds}\varphi(X_s)=\langle \nabla\varphi (X_s), U_s^\star \rangle.
\end{align}
By Equation~\ref{eq:u-star-bound} and the boundedness of $\nabla\phi$, the right side of Equation~\ref{eq:chain-rule-deriv} is bounded uniformly over $s\in[0,1]$.
Applying the fundamental theorem of calculus and Fubini's theorem to Equation~\ref{eq:chain-rule-deriv} yields
\begin{align}
    \int\varphi\ d\rho_b -\int\varphi\ d\rho_a &= \mathbb{E}[\varphi(X_b)-\varphi (X_a)\mid C,\mathcal{K}] \nonumber \\ 
    &=\int_a^b \mathbb{E} [\langle \nabla\varphi (X_s),U_s^\star \rangle \mid C,\mathcal{K}]\ ds \nonumber \\
    &=\int_a^b\int_{\mathcal{X}_{\mathcal{G}}^0} \langle \nabla\varphi (x),v^\star (x,s,C,\mathcal{K}) \rangle \rho_s(dx)\ ds, \label{eq:integral-identity} 
\end{align}
for any $0\leq a \leq b \leq 1$, where the last equality follows from conditioning on $X_s$.
The integral identity in Equation~\ref{eq:integral-identity} is the weak form of the continuity equation driven by the velocity field $v^\star$.
The initial and terminal measures are specified by the source and target distributions, respectively.
Namely,
\begin{align*}
    \rho_0 = q_0 (\cdot \mid C,\mathcal{K}), \quad \rho_1 = p_Y (\cdot\mid C,\mathcal{K}).
\end{align*}
Select a measurable velocity field $v^\star$ such that $v^\star(X_s,s,C,\mathcal{K}) =\mathbb{E}[U_s^\star \mid X_s,s,C;\mathcal{K}]$ almost surely. 
Set $v^\star=0$ on exceptional sets under the regression distribution.
Then, $v^\star$ satisfies $\lVert v^\star (X_s,s,C,\mathcal{K}) \rVert\leq \pi \sqrt{(J-1)T_{\mathrm{f}}}$ for every $x\in\mathcal X_{\mathcal G}^{0}$ and $s\in[0,1]$.
Suppose there exists a measurable map $\Phi_{s,0}:\mathcal{X}_{\mathcal{G}}^0\to\mathcal{X}_{\mathcal{G}}^0$ such that $s\mapsto\Phi_{s,0}(z)$ is absolutely continuous for almost every initial condition $z$ with respect to $q_0(\cdot\mid C,\mathcal K)$. 
For these initial conditions, the differential equation
\begin{align}
    \Phi_{0,0}(z)=z,\quad \frac{d}{ds}\Phi_{s,0}(z)=v^\star (\Phi_{s,0}(z),s,C,\mathcal{K})
\end{align}
holds for almost every $s\in[0,1]$.
Then, suppose that the weak continuity equation in Equation~\ref{eq:integral-identity} has a unique, narrowly continuous probability solution with initial distribution $q_0(\cdot\mid C,\mathcal{K})$. 
Narrow continuity means that integrating any continuous function against the probability distribution yields a continuous function in $s$.
Let $\hat{\rho}_s$ denote the distribution of $\Phi_{s,0}(Z)$, where $Z\sim q_0 (\cdot\mid C,\mathcal{K})$. 
Applying the chain rule along $\Phi_{s,0}$, integrating from $a$ to $b$, taking the conditional expectation given $(C,\mathcal{K})$, and applying the definition of $\hat{\rho}_s$ yields
\begin{align*}
    \int\varphi\,d\widehat\rho_b-\int\varphi\,d\widehat\rho_a
    =\int_a^b\int_{\mathcal X_{\mathcal G}^{0}}\left\langle\nabla\varphi(x),v^\star(x,s,C;\mathcal K)\right\rangle\widehat\rho_s(dx)\,ds.
\end{align*}
Therefore, $\hat{\rho}_s$ satisfies the same weak continuity equation and initial condition as $\rho_s$.
%
%
Both curves are narrowly continuous because their trajectories are continuous and $\mathcal X_{\mathcal G}^{0}$ is compact.
The assumed uniqueness of the weak continuity equation with the given initial distribution therefore yields
\begin{align*}
\hat\rho_s=\rho_s,\qquad s\in[0,1],
\end{align*}
and, in particular, $\hat\rho_1=\rho_1=p_Y(\cdot\mid C,\mathcal K)$.

\subsection{Preservation of Input Bone Lengths}
By Equation~\ref{eq:source}, the input bone direction vectors form a unit norm.
Therefore, their norms remain equal to $1$ throughout their trajectory as physical time progresses.
Let $\widehat{X}_s$ be a solution of Equation~\ref{eq:gsfm_dynamics}.
The projection of the velocity field onto the tangent space in Equation~\ref{eq:tangent_readout} preserves the squared norm of each bone direction vector. 
Therefore,
\begin{align*}
    \frac{d}{ds}\|\widehat X_{s,t,j}\|^2
    =2\left\langle\widehat X_{s,t,j},v_\theta(\widehat X_s,s,C;\mathcal K)_{t,j}\right\rangle
    =0,\quad j\in V\setminus\{0\}.
\end{align*}
Then, for each generation time $s\in [0,1]$, let $p_{s,t,j}$ denote the Cartesian position reconstructed from $\widehat{X}_{s,t,j}$.
Substituting $\widehat{X}_{s,t,j}$ into Equation~\ref{eq:forward_kinematics} yields
\begin{align*}
    \lVert p_{s,t,j}-p_{s,t,\mathsf p(j)}\rVert =\lVert \ell_j\widehat X_{s,t,j}\rVert=\ell_j.
\end{align*}
The exponential map also preserves unit bone direction vectors. Fix $p\in\mathbb{S}^2$ and $u\neq 0\in\mathcal{T}_p\mathbb{S}^2$. Applying the identities $\lVert p\rVert =1$ and $\langle p,u \rangle=0$ gives
\begin{align*}
    \|\mathbf{exp}_p(u)\|^2
    &=\left\|\cos(\lVert u \rVert)p+\frac{\sin(\lVert u \rVert)}{\lVert u \rVert}u\right\|^2\\
    &=\cos^2(\lVert u \rVert)\|p\|^2+2\cos(\lVert u \rVert)\frac{\sin(\lVert u \rVert)}{\lVert u \rVert}\langle p,u\rangle+\frac{\sin^2(\lVert u \rVert)}{\lVert u \rVert^2}\|u\|^2\\
    &=\cos^2(\lVert u \rVert)+\sin^2(\lVert u \rVert)=1.
\end{align*}
For $u=0$, the same conclusion follows from the identity $\mathbf{exp}_p(0)=p$.

\section{Network Architecture}
\label{app:network_architecture}
\begin{figure}[t]
\centering
\includegraphics[width=\linewidth]{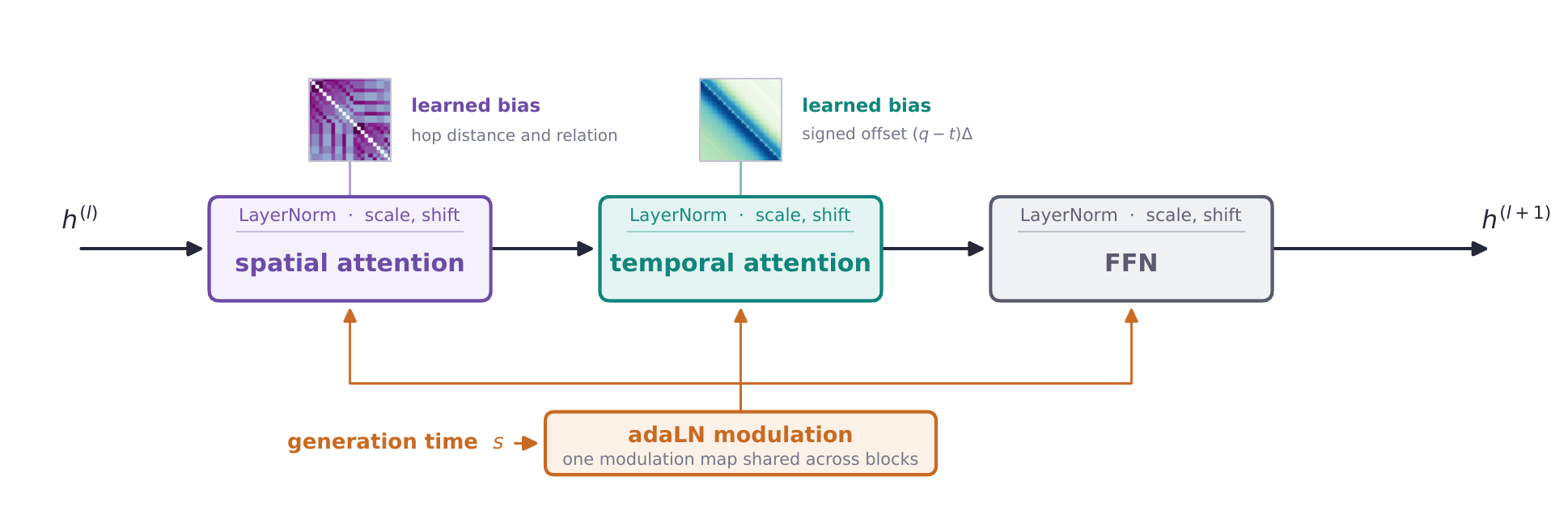}
\caption{A block of the graph structured velocity network. For visual clarity, we omit residual connections and gates.}
\label{fig:velocity_block}
\end{figure}

In this appendix, we establish the architecture of the GSFM model which produces $a_\theta(X,s,C;\mathcal K)_{t,j}$. Recall that this neural network output is projected onto the tangent space using Equation~\ref{eq:tangent_readout}. A visual representation of each GSFM block can be found in Figure~\ref{fig:velocity_block}.

\subsection{Node Features}

In this section, we describe the node features which serve as input to the GSFM model. We begin with the combined joint trajectory. The observed past motion and candidate future trajectory are joined at the physical time index $t=0$.
For each $(t,j)\in\mathcal V$, we define the combined state $\widetilde X_{t,j}$ as
\begin{align*}
    \widetilde X_{t,j}\triangleq
    \begin{cases}
    C_{t,j}, & (t,j)\in\mathcal V_{\mathrm p},\\
    X_{t,j}, & (t,j)\in\mathcal V_{\mathrm f}.
    \end{cases}
\end{align*}
The Cartesian positions $p_{t,j}\in\mathbb{R}^3$ are obtained by applying Equation~\ref{eq:forward_kinematics} to $\widetilde X$.
The observation indicator is $\beta_t\triangleq1$ for $t\in\mathcal I_{\mathrm p}$ and $\beta_t\triangleq0$ for $t\in\mathcal I_{\mathrm f}$, and denotes whether a joint state is an observed state or a future state.
The length feature $\ell_j$ denotes the length of the bone which connects the parent joint to the current joint. The length of a root node is $\ell_0\triangleq0$, since it does not have a single parent joint.
The position difference feature $\delta p_{t,j}$ denotes how far a specific joint moves between consecutive physical frames in Euclidean space.
Specifically,
\begin{align*}
    \delta p_{t,j}\triangleq
    \begin{cases}
    0, & t=-T_{\mathrm p}+1,\\
    p_{t,j}-p_{t-1,j}, & t>-T_{\mathrm p}+1.
    \end{cases}
\end{align*}
We then construct a concatenated feature $\chi_j\in\mathbb R^4$ which contains the joint depth, number of children, leaf indicator, and a root indicator.
The joint depth feature encodes the number of skeletal edges between the current joint and the root joint.
The leaf indicator is set to one if the input joint has no children, indicating that the current joint is a terminal joint.

In the subsequent feature definitions, let $D, H\in\mathbb{N}_+$ denote the time encoding dimension and feature width, respectively.
The encodings $e_t,e_s:\mathbb R\to\mathbb R^D$ represent physical time and generation time.
We use the learned map $\phi_{\mathrm{in}}:\mathbb R^{2D+15}\to\mathbb R^H$ 
\begin{align*}
    h^{(0)}_{t,j}\triangleq\phi_{\mathrm{in}}\!\left([\widetilde X_{t,j},e_t(t\Delta),\beta_t,\ell_j,e_s(s),p_{t,j},\delta p_{t,j},\chi_j]\right),
\end{align*}
to produce the initial features, where $\phi_{\mathrm{in}}$ is shared across all nodes of the GSFM model.
The input map is composed of two affine transformations and an intermediate SiLU activation.
The features of the observed states are updated in tandem with the features of future states.

\subsection{Block Structure}
The GSFM network is composed of $L\in\mathbb{N}_+$ blocks (visualized in Figure~\ref{fig:velocity_block}), which we index by $l=0,\ldots,L-1$.
The input to block $l$ is denoted by $h^{(l)}$.
Each block applies spatial attention, temporal attention, and then a feedforward network (FFN).
All operations within the block use adaptive layer normalization and a gated residual connection.

Within each block, we use a learned map $\phi_c:\mathbb R^D\to\mathbb R^H$ to produce an embedding of the generation time $c_s\triangleq\phi_c(e_s(s))$.
Additionally, we apply a shared modulation map $M:\mathbb R^H\to\mathbb R^{9H}$ to the block input, where the modulation map consists of a SiLU activation followed by an affine transformation.
In the GSFM-deep configuration, a learned offset $\omega_l\in\mathbb R^{9H}$ is added for all blocks $l=0,\ldots, L-1$.
The vector $M(c_s)+\omega_l$ contains a shift, scale, and residual gate for each of the spatial attention, temporal attention, and feedforward operations, respectively.

For a sublayer input array $h$ (i.e., inter-layer input within a single block $l$), let $F$ denote the corresponding attention or FFN transformation.
The modulation vectors associated with $h$ are the shift vector $\delta\in\mathbb R^H$, scale vector $\gamma\in\mathbb R^H$, and residual connection vector $g\in\mathbb R^H$.
Let the symbol $\mathsf{LN}$ denote layer normalization along the feature dimension, $\odot$ denote the Hadamard product, and $\mathbf1\in\mathbb R^H$ denote the vector of ones in $\mathbb{R}^H$.
The modulated input $\widetilde h$ and updated features $h^+$ are given as
\begin{align*}
    \widetilde h=(\mathbf1+\gamma)\odot\mathsf{LN}(h)+\delta,\quad h^+= h+g\odot F(\widetilde h).
\end{align*}
The modulation vectors are then broadcast over all nodes.
The FFN is applied to each node separately and consists of two affine transformations with intermediate width $4H$ followed by a GELU activation.

\subsection{Spatial Attention and Temporal Attention}
Consider one spatial attention layer having an attention head with dimension $d_{\mathrm h}\in\mathbb N_+$.
We suppress the indices in the subsequent expressions for notational clarity.
Let $\mathsf{hop}(j,i)$ denote the skeletal distance map, which returns the shortest path length between joints $j$ and $i$ in the undirected skeleton tree.
Using $\mathsf{hop}(j,i)$ and a fixed query node $j$, we construct `buckets' of nodes, each consisting of the collection of nodes which are $d\in\mathbb{N}_+$ hops away from $j$.
Let $h_{\max}\in\mathbb{N}_+$ denote the largest distance bucket considered for every query node.
Let the relation map $\mathsf{rel}(j,i)$ indicate whether joint $j$ is either: (i) identical to joint $i$, (ii) the parent of joint $i$, (iii) a child of $i$, (iv) a sibling of $i$, or (v) none of these.
A learned linear map of the modulated input features is used to produce the query, key, and value vectors $\mathbf q^{\mathrm{sp}}_{t,j},\mathbf k^{\mathrm{sp}}_{t,j},\mathbf z^{\mathrm{sp}}_{t,j}\in\mathbb R^{d_{\mathrm h}}$.
The learned scalar values $b_{\mathrm{hop}}$ and $b_{\mathrm{rel}}$ then provide biases for each of the distance buckets and relation codes, respectively.
Using the constructions above, the spatial attention score is defined as
\begin{align*}
    a^{\mathrm{sp}}_{t,j,i}\triangleq\frac{\langle\mathbf q^{\mathrm{sp}}_{t,j},\mathbf k^{\mathrm{sp}}_{t,i}\rangle}{\sqrt{d_{\mathrm h}}}+b_{\mathrm{hop}}\!\left(\min\{\mathsf{hop}(j,i),h_{\max}\}\right)+b_{\mathrm{rel}}\!\left(\mathsf{rel}(j,i)\right).
\end{align*}
The spatial attention head output is a weighted sum of the value vectors, where
\begin{align*}
    m^{\mathrm{sp}}_{t,j}\triangleq\sum_{i\in V}\frac{\exp(a^{\mathrm{sp}}_{t,j,i})}{\sum_{k\in V}\exp(a^{\mathrm{sp}}_{t,j,k})}\mathbf z^{\mathrm{sp}}_{t,i}.
\end{align*}
%
%
%
The temporal attention layer acts analogously to the spatial attention layer.
However, rather than operating on buckets of joints, temporal attention is scaled by the inter-frame distance. 
As above, we use learned linear maps having modulated temporal features as input to generate the query, key, and value vectors $\mathbf q^{\mathrm{tm}}_{t,j},\mathbf k^{\mathrm{tm}}_{t,j},\mathbf z^{\mathrm{tm}}_{t,j}\in\mathbb R^{d_{\mathrm h}}$.
We then use a learned bias network to construct a scalar encoding of the signed physical time offset to for each attention head.
For the head considered in this example, we denote the bias network by $b_{\mathrm{tm}}:\mathbb R\to\mathbb R$.
Using the constructions above, given a query frame $t$ and key frame $q$, the temporal attention score is given as
\begin{align*}
    a^{\mathrm{tm}}_{j,t,q}\triangleq\frac{\langle\mathbf q^{\mathrm{tm}}_{t,j},\mathbf k^{\mathrm{tm}}_{q,j}\rangle}{\sqrt{d_{\mathrm h}}}+b_{\mathrm{tm}}\!\left((q-t)\Delta\right).
\end{align*}
The temporal attention head output is the weighted sum of the value vectors, where
\begin{align*}
    m^{\mathrm{tm}}_{t,j}\triangleq\sum_{q\in\mathcal I}\frac{\exp(a^{\mathrm{tm}}_{j,t,q})}{\sum_{u\in\mathcal I}\exp(a^{\mathrm{tm}}_{j,t,u})}\mathbf z^{\mathrm{tm}}_{q,j}.
\end{align*}
%
%
%
The spatial and temporal bias parameters are shared across blocks in both the GSFM-deep and GSFM-tied configurations.

\subsection{Output and Parameter Sharing}
Given an input $(X,s,C,\mathcal K)$, let the notation $a_{t,j}$ serve as shorthand for the network output $a_\theta(X,s,C;\mathcal K)_{t,j}$ before its projection onto the tangent space, as in Equation~\ref{eq:tangent_readout}.
Next, given $c_s$ as input (which we recall is the output of the generation time embedding map $\phi_c$ having as input $e_s(s)$), we apply a SiLU activation and an affine transformation to produce the readout shift and scale vectors $\delta_s^{\mathrm{out}},\gamma_s^{\mathrm{out}}\in\mathbb R^H$.
A shared affine map $A_{\mathrm{out}}:\mathbb R^H\to\mathbb R^3$ is then used to produce the ambient future outputs
\begin{align*}
    a_{t,j}\triangleq A_{\mathrm{out}}\!\left((\mathbf1+\gamma_s^{\mathrm{out}})\odot\mathsf{LN}(h^{(L)}_{t,j})+\delta_s^{\mathrm{out}}\right),\quad (t,j)\in\mathcal V_{\mathrm f}.
\end{align*}
Lastly, using $a_{t,j}$, the projection onto the tangent space is given by Equation~\ref{eq:tangent_readout}.
%
%

The GSFM-deep configuration uses distinct attention and FFN parameters for each block $l=0,1,\ldots, L-1$.
The GSFM-tied configuration applies a single shared block $L$ times.
At each block $l$, a learned vector $\iota_l\in\mathbb R^H$ is used to modify the conditioning to $c_s+\iota_l$, yielding the modulation vector $M(c_s+\iota_l)+\omega$. 
Recall that $\omega\in\mathbb R^{9H}$ denotes the shared block offset.
%
%
In Table~\ref{tab:architecture_configuration}, we describe the structural parameters used in the GSFM-deep and GSFM-tied experiments.

\begin{table}[t]
\centering
\caption{Structural parameters for the GSFM-deep and GSFM-tied configurations.}
\label{tab:architecture_configuration}
\small
\begin{tabular}{lcc}
\toprule
Setting & Deep & Tied \\
\midrule
Feature width $H$ & $384$ & $384$ \\
Block applications $L$ & $12$ & $12$ \\
Attention heads & $8$ & $8$ \\
Time encoding dimension $D$ & $128$ & $128$ \\
Attention and FFN parameters across depth & Distinct & Shared \\
Parameter count in millions & $30.49$ & $4.47$ \\
\bottomrule
\end{tabular}
\end{table}

\section{Numerical Implementation and Training Objective}
\label{app:numerical_implementation}
In this appendix, we describe the numerical operations, implemented loss function, and integration procedure used in Algorithms~\ref{alg:gsfm_training} and~\ref{alg:gsfm_generation}.

\subsection{Exponential Updates and Transport}
First, we define the norm floor map $\nu:\mathbb{R}^3\to\mathbb{R}_{>0}$ and the normalization map $\mathcal{R}:\mathbb{R}^3\to\mathbb{R}^3$. The norm floor and normalization maps are given as
\begin{align*}
    \nu(w)\triangleq\sqrt{\max\{\|w\|^2,\varepsilon_{\mathrm g}^2\}},\quad \mathcal R(w)\triangleq\frac{w}{\nu(w)},
\end{align*}
respectively, where $\varepsilon_{\mathrm g}$ is a user-selected constant denoting the lower bound of the norm floor. In all numerical experiments, we use $\varepsilon_{\mathrm g}\triangleq10^{-6}$.
The map $\mathcal R$ returns a unit vector whenever $\|w\|\geq\varepsilon_{\mathrm g}$.
Given $p\in\mathbb S^2$ and $u\in\mathcal T_p\mathbb S^2$, the numerical implementation of the exponential map is defined as
\begin{align*}
    \mathbf{exp}^{\mathrm{num}}_p(u)\triangleq\mathcal R\!\left(\cos(\nu(u))p+\frac{\sin(\nu(u))}{\nu(u)}u\right).
\end{align*}
We use the numerical exponential map for all integration updates and to construct the source.

For $p,q\in\mathbb S^2$ and $u\in\mathcal T_p\mathbb S^2$, the numerical implementation of parallel transport is given as:
\begin{align*}
    \mathcal T^{\mathrm{num}}_{p\to q}(u)\triangleq
    \begin{cases}
    u-\dfrac{\langle q,u\rangle}{1+\langle p,q\rangle}(p+q), & |1+\langle p,q\rangle|>\varepsilon_{\mathrm g},\\
    \Pi_q(u), & |1+\langle p,q\rangle|\leq\varepsilon_{\mathrm g}.
    \end{cases}
\end{align*}
For Euclidean input values, $\mathbf{exp}^{\mathrm{num}}$ is addition and $\mathcal T^{\mathrm{num}}$ is the identity.
On the product state space, the exponential map and parallel transport are applied separately to each input node.
When given a trajectory as input, $\mathcal R$ is applied to the bone components and leaves the root state unchanged.

\subsection{Numerical Training Paths}
Given bone endpoints $y,z\in\mathbb S^2$, denote their inner product by $c$, the clipped angle map by $\alpha$, and the projected direction by $b$ such that
\begin{align*}
    c&\triangleq\langle z,y\rangle,\\
    \alpha&\triangleq\arccos\!\left(\min\{1-\varepsilon_{\mathrm g},\max\{-1+\varepsilon_{\mathrm g},c\}\}\right),\\
    b&\triangleq y-cz.
\end{align*}
The normalized projected direction is given as $a\triangleq b/\nu(b)$.
For $s\in[0,1]$, the corresponding point on the path $q_s^{\mathrm{num}}$ and target velocity $u_s^{\mathrm{num}}$ are defined as
\begin{align}
    q_s^{\mathrm{num}}&\triangleq
    \begin{cases}
    z, & \|b\|<\varepsilon_{\mathrm g},\\
    \mathcal R\!\left(\cos(s\alpha)z+\sin(s\alpha)a\right), & \|b\|\geq\varepsilon_{\mathrm g},
    \end{cases}
    \label{eq:numerical_path} \\
    u_s^{\mathrm{num}}&\triangleq
    \begin{cases}
    0, & \|b\|<\varepsilon_{\mathrm g},\\
    \alpha\big(-\sin(s\alpha)z+\cos(s\alpha)a\big), & \|b\|\geq\varepsilon_{\mathrm g}.
    \end{cases}
    \label{eq:numerical_target}
\end{align}
The numerical product path is computed for every input bone specified by two connected joint nodes.
Throughout generation, the root position and root target velocity is set to zero.
%

\subsection{Source Samples and Batch Loss}
Training is performed in batches. To construct a training batch, we assemble $m\in\mathbb N_+$ samples $(C_i,Y_i,\mathcal K_i)$ indexed by $i=1,\ldots,m$.
For every sample, independently draw $\xi_{i,t,j}\sim\mathcal N(0,I_3)$.
Using $\xi_{i,t,j}$, we construct the source numerically as
\begin{align}
    Z_{i,t,0}\triangleq0,\quad Z_{i,t,j}\triangleq\mathbf{exp}^{\mathrm{num}}_{C_{i,0,j}}\!\left(\sigma_{\mathrm d}\Pi_{C_{i,0,j}}(\xi_{i,t,j})\right),\quad j\in V\setminus\{0\}.
    \label{eq:numerical_source}
\end{align}
We use independently drawn scalars $\zeta_i\sim\mathcal N(0,1)$ to determine the generation time $s_i\triangleq(1+\exp(-\zeta_i))^{-1}$ for each sample.
The resulting time distribution has positive density on $(0,1)$.
Applying Equation~\ref{eq:numerical_path} and Equation~\ref{eq:numerical_target} to $(Z_i,Y_i)$ at $s_i$ gives the trajectory $Q_i$ and target velocity $U_i$.

We then iterate over the samples to construct the set of retained nodes $\mathcal A$ as
\begin{align*}
    \mathcal A\triangleq\left\{(i,t,j)\in\{1,\ldots,m\}\times\mathcal I_{\mathrm f}\times(V\setminus\{0\}):\langle Z_{i,t,j},Y_{i,t,j}\rangle\geq-1+\varepsilon_{\mathrm{cut}}\right\},
\end{align*}
where $\varepsilon_{\mathrm{cut}}$ is a threshold used to exclude nodes whose observed future state is close to being antipodal with its source.
In all experiments, we use $\varepsilon_{\mathrm{cut}}\triangleq10^{-4}$.
We use the per-batch loss function
\begin{align}
    \widehat{\mathcal L}_{\mathrm{FM}}(\theta)\triangleq\frac{\displaystyle\sum_{(i,t,j)\in\mathcal A}\left\|v_\theta(Q_i,s_i,C_i;\mathcal K_i)_{t,j}-U_{i,t,j}\right\|^2}{\max\{|\mathcal A|,1\}}.
    \label{eq:training_loss}
\end{align}
The denominator scales the loss according to the number of retained nodes.
Training minimizes its expectation through stochastic parameter updates.
This objective differs from Equation~\ref{eq:ideal_fm_loss} through (i) the numerical implementation of the source and path and (ii) the exclusion of a neighborhood about the antipodal configuration.

\subsection{Optimization and Sampling Settings}
To facilitate training, we use AdamW with a base learning rate of $2\times10^{-4}$, moment coefficients of $(0.9,0.95)$, a weight decay of $0.01$ on matrix parameters, and gradient clipping with a threshold of $1.0$.
The schedule uses $2{,}000$ warmup updates followed by cosine decay to $2\%$ of the base learning rate.
The full AMASS model configurations use a total of $94{,}200$ training updates with a batch size of $64$ samples per device on four NVIDIA B200 GPUs.

We maintain a running average of the parameters $\theta$ during training updated as 
\begin{align*}
\bar\theta\gets\beta_{\mathrm{ema}}\bar\theta+(1-\beta_{\mathrm{ema}})\theta,
\end{align*}
where $\beta_{\mathrm{ema}}$ denotes a decay rate.
In all experiments, we use a decay rate of $\beta_{\mathrm{ema}}\triangleq0.999$.
The final model which is used for evaluation has parameters $\bar{\theta}$.
Unless stated otherwise, the source scale is $\sigma_{\mathrm d}=0.7$ during both training and generation, and sampling uses $32$ bit floating point arithmetic and $25$ midpoint steps.
Each midpoint step requires two velocity field evaluations, resulting in a total of $50$ evaluations for the experiment results.

\section{Experimental Protocols}
\label{app:experimental_protocols}
In this appendix, we specify the data protocols, metric reductions, and comparison settings used in Section~\ref{sec:experiments}.
The GSFM-deep and GSFM-tied network configurations and optimization settings are provided in Appendices~\ref{app:network_architecture} and~\ref{app:numerical_implementation}, respectively.

\subsection{Data and Prediction Task}
The skeleton provided in the AMASS dataset has $22$ joints, which is used to construct the kinematic tree.
For all experiments, we use the BeLFusion partition of $11$ training, $4$ validation, and $7$ test datasets~\citep{mahmood2019amass,barquero2023belfusion}.
AMASS evaluation uses $12{,}742$ test segments under the published segmentation protocol.
%
%
The skeleton provided in the Human3.6M dataset has $17$ joints.
We resample the Human3.6M data from $50$ to $60$ frames per second to match the frame rate of AMASS~\citep{ionescu2014human36m}.
The performance evaluation on the Human3.6M dataset contains $5{,}169$ segments from subjects S9 and S11.
The GSFM models used for this zero-shot evaluation are trained on AMASS.
The parameters of the evaluated models are not updated before testing on Human3.6M.

%
Each input test observation yields $50$ generated future trajectories, which are then used to compute error metrics.
%
%
The spatial attention biases are evaluated using the target skeletal kinematic tree, while the learned parameters and the source scale remain fixed.
The baseline entries in Tables~\ref{tab:amass_results} and~\ref{tab:zero_shot} are reproduced from EquiFusion~\citep{curreli2026equifusion}.

\subsection{Metrics and Reporting Conventions}

\textbf{Prediction errors.}
The average displacement error (ADE) and final displacement error (FDE) evaluate all $50$ generated future trajectories against a single reference future trajectory and return the the minimum mean trajectory error and minimum terminal error, respectively.
The multimodal ADE and FDE (MMADE and MMFDE) evaluate all $50$ generated future trajectories against a collection of reference future trajectories, and return the mean of the ADE and FDE with respect to the collection of references.
The mean angle error (MAE) compares the angles between adjacent limbs along each kinematic chain and averages the absolute angular errors over limb pairs and future frames.
We report its minimum over the 50 samples.

\textbf{Diversity and motion statistics.}
The average pairwise distance (APD) is the mean Euclidean distance between pairs of generated future trajectories after flattening the frame, joint, and coordinate dimensions. APD allows us to quantify variation between generated future trajectories. The average pairwise distance error (APDE) is the absolute difference between the APD of the generated futures and the APD of the corresponding multimodal reference set.
The cumulative motion distribution error (CMD) measures deviations between generated displacement statistics and the expected reference motion statistics.

\textbf{Skeletal consistency.}
Limb stretching (Str.) and limb jitter (Jit.) measure bone length inconsistency across a trajectory and its temporal variation, respectively.
Both Str. and Jit. are expressed as percentages.
We note that lower values are preferred for all error metrics.

\textbf{Metric reductions.}
Tables~\ref{tab:amass_results}, \ref{tab:zero_shot}, \ref{tab:parameter_sharing}, and~\ref{tab:inference_budget} leverage the benchmark metric reductions of~\cite{curreli2026equifusion}.
%
%
At each frame, ADE and FDE use the Euclidean norm of the flattened pose difference, with joint coordinates expressed in meters.
These quantities are not mean distances over individual joints.
Table~\ref{tab:graph_ablations} instead uses ADE and FDE that average joint distances.
Its APD divides each pairwise trajectory distance by the square root of the number of frame and joint pairs.
The ADE, FDE, and APD values in Table~\ref{tab:graph_ablations} are reported in centimeters.
These differences prevent direct comparison with the corresponding benchmark entries.
%

\subsection{Spatial and Temporal Attention Ablations}
The `full spatial and temporal attention' variation in the ablation study has a hidden width of $256$, $6$ GSFM blocks, and a generation-time modulation map which is shared across blocks.
Every configuration considered in the ablation study is trained for $4{,}000$ updates and evaluated on $600$ AMASS segments.
To evaluate each generation, we generate $50$ future trajectories per input observed trajectory.
Let NFE refer to the number of function evaluations.
Specifically, NFE refers to the number of times the velocity field was constructed numerically for a single generated trajectory.
Each generated trajectory has $10$ midpoint steps, resulting in a NFE of $20$.
We trained each variation in the ablation study over two seeds.
We then collected result for each seed.
The reported results represent the mean over both seeds.

Spatial one hop attention computes attention weights using the current joint node and its directly connected skeletal neighbors.
Temporal one hop attention takes the current frame and its temporally adjacent frames as input.
%
%
The `spatial one hop' configuration applies spatial attention only to the current joint and its directly connected skeletal neighbors, while temporal attention is implemented using its standard configuration.
The `temporal one hop' configuration applies temporal attention only to the current and directly adjacent frames of the same joint, while spatial attention is implemented using its standard configuration.
Both the spatial and temporal one hop restrictions have the same parameter count as the full attention configuration.
%
%
%
%
%
%

\subsection{Parameter Sharing and Source Scale}
The structural parameters which specify the GSFM-deep and GSFM-tied models are detailed in Appendix~\ref{app:network_architecture}.
The two GSFM configurations use the same depth of 12 blocks and are compared using source scale values $\sigma_d\in\{0.5,0.7,0.9\}$.
The primary experimental configuration for both GSFM-deep and GSFM-tied uses $\sigma_d=0.7$.
%
%
We note that the GSFM-deep and GSFM-tied models in Table~\ref{tab:parameter_sharing} are trained separately at $\sigma_{\mathrm d}\in\{0.3,0.5,0.7,0.9\}$ and $\sigma_{\mathrm d}\in\{0.5,0.7,0.9\}$, respectively.
Each model uses the same source scale value from training during generation, since changing the source scale alters the source distribution in Equation~\ref{eq:source} and the associated training paths.
Both the GSFM-deep and GSFM-tied configurations are trained using a single run, so the reported results are not presented as statistically significant.
%
%

\section{Inference Cost}
\label{app:inference_cost}
Table~\ref{tab:inference_budget} demonstrates the sampling cost of the GSFM-deep model on a single NVIDIA B200 GPU using $32$ bit floating point arithmetic.
To conduct a sampling call using GSFM-deep or GSFM-tied, $50$ future trajectories are generated for a single input observation.
Each generated trajectory is made up of $120$ future frames.
%
%
The output generation timing accounts for source sampling and all numerical integration steps, but excludes data loading, the conversion to Cartesian joint positions, and evaluation of performance metrics.
%

\begin{table*}[t]
\centering
\caption{Sampling cost of GSFM-deep on AMASS for a single input observation and $50$ generated predicted future trajectories having $120$ frames each. Measurements use one NVIDIA B200 GPU and $32$ bit floating point precision. The total time accounts for source sampling and numerical integration steps, but excludes data loading, conversion to Cartesian joint positions, and evaluation of performance metrics. Peak memory is reported in mebibytes (MiB).}
\label{tab:inference_budget}
\setlength{\tabcolsep}{4pt}
\small
\resizebox{\textwidth}{!}{%
\begin{tabular}{cccccccc}
\toprule
\multicolumn{2}{c}{Sampling}
& \multicolumn{2}{c}{Cost}
& \multicolumn{2}{c}{Prediction}
& \multicolumn{1}{c}{Diversity}
& \multicolumn{1}{c}{Motion statistics} \\
\cmidrule(lr){1-2}
\cmidrule(lr){3-4}
\cmidrule(lr){5-6}
\cmidrule(lr){7-7}
\cmidrule(lr){8-8}
Midpoint steps
& NFE
& Total time (s)
& Peak memory (MiB)
& ADE $\downarrow$
& FDE $\downarrow$
& APDE $\downarrow$
& CMD $\downarrow$ \\
\midrule
\multicolumn{8}{l}{\textit{Inference performance at $\sigma_{\mathrm d}=0.9$}} \\
25 & 50 & 15.659 & 3,533 & 0.4838 & 0.5522 & 2.513 & 4.964 \\
10 & 20 & 6.261 & 3,533 & 0.4904 & 0.5556 & 2.117 & 4.928 \\
\midrule
\multicolumn{8}{l}{\textit{Inference performance at $\sigma_{\mathrm d}=0.7$}} \\
25 & 50 & 15.659 & 3{,}533 & 0.4775 & 0.5505 & 2.840 & 5.614 \\
10 & 20 & 6.264 & 3{,}533 & 0.4787 & 0.5520 & 2.886 & 5.926 \\
\bottomrule
\end{tabular}}
\end{table*}

For the source scale used in the main numerical experiments ($\sigma_{\mathrm d}=0.7$), reducing the integration budget from 50 to 20 NFE lowers results in an approximately $60\%$ lower sampling time.
Reducing the integration budget corresponds to small increases in ADE and FDE and larger errors in diversity calibration and motion statistics.
%
%

%
%
%
%
%
%

\section{Additional Related Work}
\label{app:additional_related}
\paragraph{Flow matching for motion generation.}
The generalized conditional flow matching formulation of \citet{tong2024improving} accommodates different source distributions and couplings between source and target samples.
Motion Flow Matching generates complete motion sequences conditioned on text or action labels \citep{hu2023motion}.
Motion Flow Matching also facilitates motion prediction by incorporating observed motion prefixes through sampling trajectory rewriting.
CacheFlow trains an unconditional continuous normalizing flow using flow matching and caches latent representations together with their associated motions \citep{maeda2025cacheflow}.
A conditional Gaussian mixture inferred from observed motion enables prediction using these cached results.
Our developed GSFM model advances the techniques used for human motion prediction through graph structured conditional transport on a product of unit spheres.
In our method, a single tangent velocity field $v_\theta$ jointly evolves all future bone directions and depends explicitly on both (i) the observed motion, and (ii) the supplied kinematic tree.
This formulation places the provided skeletal structure within the learned transport and preserves input bone lengths throughout the generation process.

\paragraph{Spatial and temporal modeling.}
\citet{aksan2021spatio} use separate spatial and temporal attention mechanisms for autoregressive human motion prediction.
\citet{sofianos2021space} factor spatiotemporal graph connectivity into learned spatial and temporal affinity matrices for pose forecasting.
MotionDiff incorporates spatial and temporal modeling into a diffusion network for stochastic motion prediction and uses a separate refinement network with geometric losses \citep{wei2023human}.
Within this context, our GSFM model couples spatial and temporal interactions within the tangent velocity field $v_\theta$ which transports the complete future skeletal trajectory.
Skeletal relations and signed physical time offsets inform every generation step, while the spherical representation and geometric integration preserve bone lengths without requiring a separate refinement network.
Our ablations demonstrate the contributions of both spatial and temporal attention to the accuracy of predicted trajectories.
On the AMASS dataset, GSFM-deep achieves the lowest ADE and MAE, while GSFM-tied achieves the lowest CMD among the methods compared in Table~\ref{tab:amass_results}.
Our results yield leading prediction accuracy and motion statistics, despite a significant reduction in parameter counts relative to the compared baselines.

\paragraph{Graph structure and generative conditioning.}
Graphormer incorporates structural information through learned attention biases indexed by the shortest path distance between nodes \citep{ying2021transformers}.
Diffusion Transformers condition transformer blocks through adaptive layer normalization, and learn modulation of residual branches \citep{peebles2023scalable}.
Our GSFM model integrates structural conditioning and generation time modulation into the tangent velocity field $v_\theta$ for complete skeletal trajectories.
The supplied kinematic tree and shared node operations allow the model to process different skeletal structures without modifying learned parameters.
Models trained on AMASS with 22 joints transfer directly to the unseen Human3.6M dataset with a different skeleton of 17 joints, without retargeting or parameter updates.
In this zero-shot evaluation, both GSFM variants (-deep and -tied) achieve lower ADE and FDE than every baseline which used retargeting in Table~\ref{tab:zero_shot}.
GSFM-deep also achieves the lowest CMD among all methods in the Table~\ref{tab:zero_shot} comparison.
The GSFM-tied variant similarly supports zero-shot transfer while having only 4.47 million parameters.
Together, these results demonstrate that GSFM combines accurate motion prediction with bone length preservation, while also supporting zero-shot transfer across datasets and skeletal structures.

\end{document}